\def\year{2021}\relax
\documentclass[letterpaper]{article} 
\usepackage{aaai21}  
\usepackage{amsmath}
\usepackage{amsthm}
\usepackage{amssymb}
\usepackage{amsfonts}
\usepackage{times}  
\usepackage{helvet} 
\usepackage{courier}  
\usepackage{algorithm}
\usepackage{algorithmic}
\usepackage{subfigure} 
\usepackage[capitalise]{cleveref}
\usepackage[hyphens]{url}  
\usepackage{graphicx} 
\def\UrlFont{\rm}  
\usepackage{natbib}  
\usepackage{caption} 
\newcommand{\ie}{\textit{i}.\textit{e}., }
\newcommand{\eg}{\textit{e}.\textit{g}., }

\title{Stochastic Adversarial Gradient Embedding for Active Domain Adaptation}
\author {

        Victor Bouvier\textsuperscript{\rm 1, 2},
        Philippe Very\textsuperscript{\rm 3},
        Clément Chastagnol\textsuperscript{\rm 4},
        Myriam Tami\textsuperscript{\rm 1},
        Céline Hudelot\textsuperscript{\rm 1} \\
}
\affiliations {
    \textsuperscript{\rm 1}  Universit\'e Paris-Saclay, CentraleSup\'elec, Math\'ematiques et Informatique pour la
Complexit\'e et les Syst\`emes, 91190, Gif-sur-Yvette, France \\
    \textsuperscript{\rm 2} Sidetrade, 114 Rue Gallieni, 92100, Boulogne-Billancourt, France \\
    \textsuperscript{\rm 3}  Lend-Rx, 24 Rue Saint Dominique, 75007, Paris, France \\
    \textsuperscript{\rm 4}  Alan, 117 Quai de Valmy, 75010 Paris, France, \\
    vbouvier@sidetrade.com, philippe.very@lend-rxtech.com, clement.chastagnol@alan.eu, firstname.name@centralesupelec.fr
}

\begin{document}

\maketitle

\begin{abstract}
Unsupervised Domain Adaptation (UDA) aims to bridge the gap between a source domain, where labelled data are available, and a target domain only represented with unlabelled data. If domain invariant representations have dramatically improved the adaptability of models, to guarantee their good transferability remains a challenging problem. This paper addresses this problem by using active learning to annotate a small budget of target data. Although this setup, called Active Domain Adaptation (ADA), deviates from UDA's standard setup, a wide range of practical applications are faced with this situation. To this purpose, we introduce \textit{Stochastic Adversarial Gradient Embedding} (SAGE), a framework that makes a triple contribution to ADA. First, we select for annotation target samples that are likely to improve the representations' transferability by measuring the variation, before and after annotation, of the transferability loss gradient. Second, we increase sampling diversity by promoting different gradient directions. Third, we introduce a novel training procedure for actively incorporating target samples when learning invariant representations. SAGE is based on solid theoretical ground and validated on various UDA benchmarks against several baselines. Our empirical investigation demonstrates that SAGE takes the best of uncertainty \textit{vs} diversity samplings and improves representations transferability substantially.
\end{abstract}

\section{Introduction}

When provided with a large amount of labelled data, deep neural networks have dramatically improved the state-of-the-art for both vision \cite{krizhevsky2012imagenet} and language tasks \cite{vaswani2017attention}. However, deploying machine learning models for real-world applications requires to generalize on data which may slightly differ with the training data \cite{amodei2016concrete, marcus2020next}. Quite surprisingly, deep models do not meet this requirement and often show a weak ability to generalize out of the training distribution \cite{beery2018recognition, geva2019we, arjovsky2019invariant}. 

Deep nets can learn data transferable representations to new tasks or new domains if some labelled data from the new distribution are available \cite{oquab2014learning, yosinski2014transferable}. Acquiring a sufficient amount of labeled data is laborious, and large scale annotation is often cost-prohibitive. Unlabelled data are much more convenient to obtain. This observation has motivated the field of \textit{Unsupervised Domain Adaptation} \cite{pan2009survey, quionero2009dataset} for bridging the gap between a labelled \textit{source domain} and an unlabelled \textit{target domain}. 

Learning domain \textit{Invariant Representations} has led to significant progress towards learning domain transferable representations with deep neural networks \cite{ganin2015unsupervised, long2015learning, long2018conditional}. By fooling a discriminator trained to separate the source from the target domain, the feature extractor removes domain-specific information in representations \cite{ganin2015unsupervised}. Therefore, a classifier trained from those representations with source labelled data is expected to perform reasonably well in the target domain \cite{ben2007analysis, ben2010theory}.

\begin{figure*}
    \centering
    \includegraphics[scale=0.3]{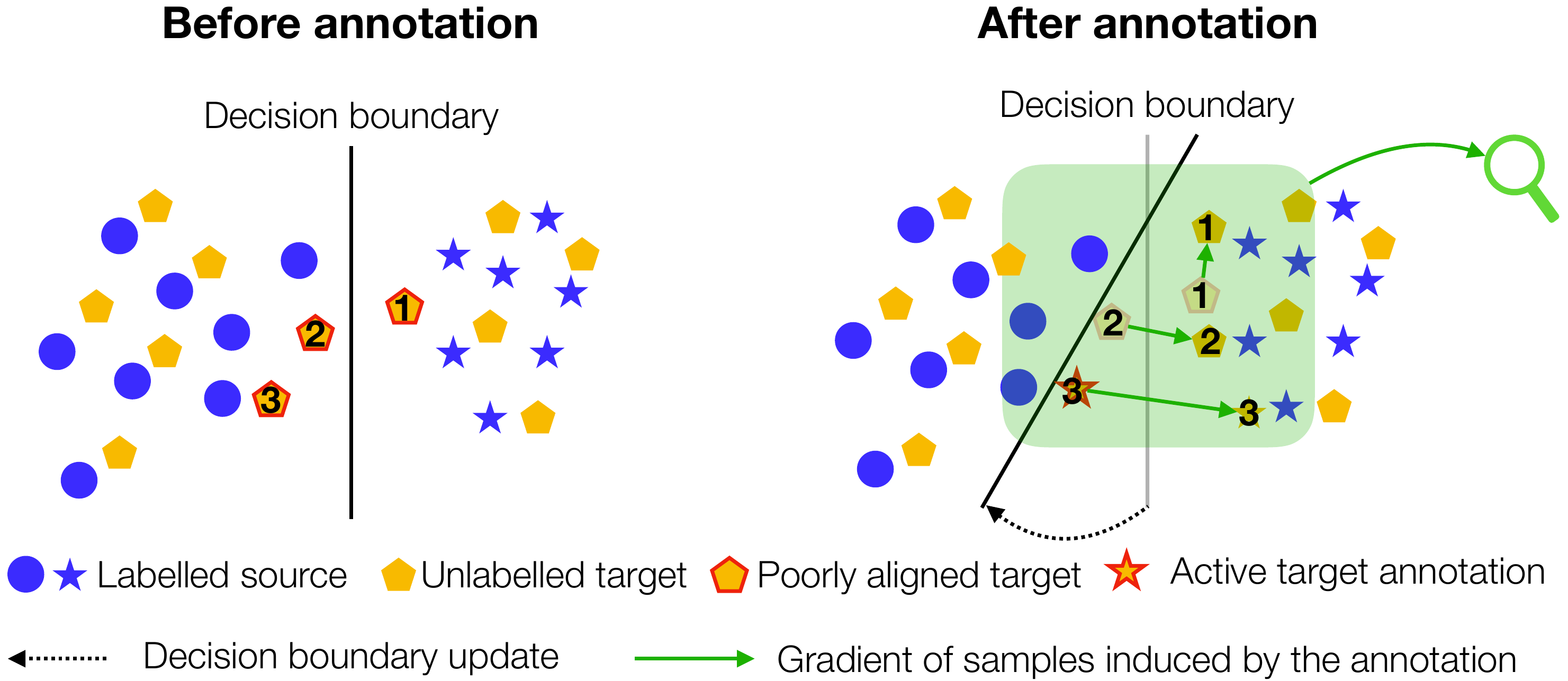} \includegraphics[scale=0.55]{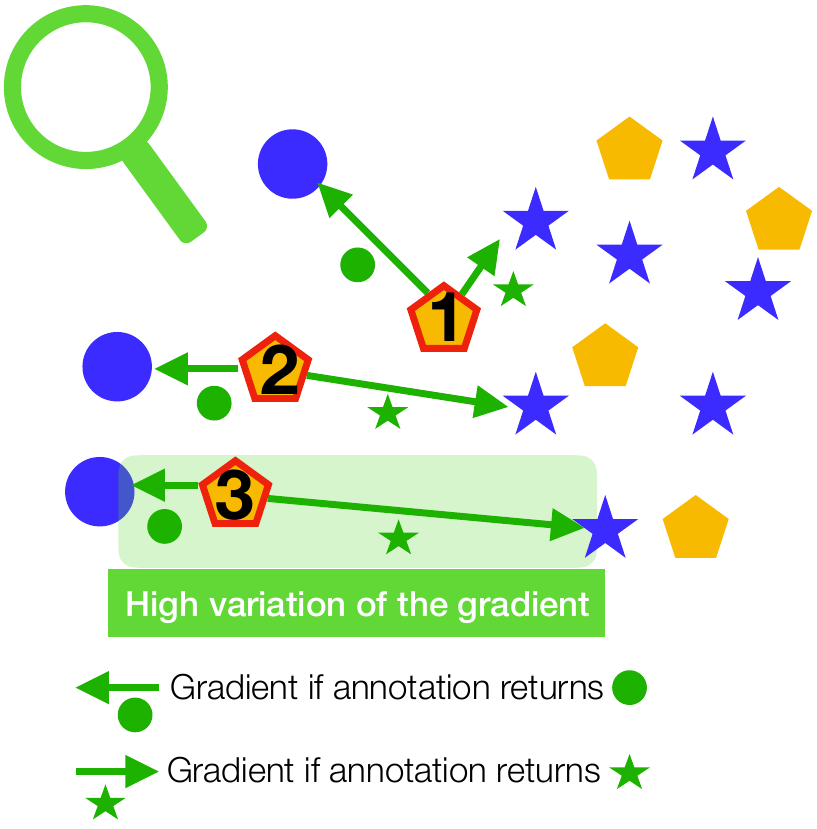}
    \caption{Illustration of the effect of annotation of a target sample selected by SAGE (\textit{best viewed in colors}). Binary classification problem ($\bullet$ \textit{vs} $\star$) where source samples are blue and target samples are orange. Before annotation, the class-level alignment is not satisfactory leading to a potential negative transfer (poorly aligned target samples tagged as \textbf{1}, \textbf{2} and \textbf{3}). We estimate which sample should be annotated as a priority by measuring the variation, before and after annotation, of the gradient of the representations' transferability. We observe the highest variation is obtained for target sample \textbf{3}, which is sent to an oracle. The oracle annotation returns class $\star$, validating the suspicion of negative transfer. This leads to an update of the decision boundary which pushes \textbf{1}, \textbf{2} and \textbf{3} into class $\star$, resulting of a better class-level alignment of representations.}
    \label{fig:active_demo}
\end{figure*}

However, those methods perform significantly worse than their fully supervised counterparts. Practical applications often offer the possibility of annotating a fixed budget of target data; a paradigm referred to as \textit{Active Learning} (AL). Despite its great practical interest, there are, to our knowledge, only a few previous works which address the problem of \textit{Active Domain Adaptation} (ADA)  \cite{chattopadhyay2013joint, rai2010domain, saha2011active, su2020active}. In particular, the work of \citeauthor{su2020active}, proposed recently, is the first that brings active learning to domain adversarial learning.

Our insight is to reserve the annotation budget for the data that will have the most impact on the representations' transferability. Promoting class-level domain invariance \cite{long2018conditional, bouvier2020robust} makes it possible to assess this impact precisely. Indeed, class-level invariance consists of using predicted target labels for soft-class conditioning in the domain adversarial loss. Involving an oracle, which provides the ground-truth, enables to change the contribution of a target sample from soft to hard-class conditioning. Therefore, we measure the impact of annotation by estimating the variation of the domain adversarial loss gradient (see \cref{fig:active_demo}). Thus, in this paper, 

\begin{itemize}
    \item We present \textbf{S}tochastic \textbf{A}dversarial \textbf{G}radient \textbf{E}mbedding (SAGE), an embedding of target samples suited for active learning of domain invariant representations. This embedding is obtained by measuring the variation, before and after annotation, of the domain adversarial loss gradient. SAGE has the following properties: 
    \begin{enumerate}
        \item It takes into account the uncertainty on a target sample label by considering the gradient as a stochastic vector. Since annotation removes uncertainty, it allows us to estimate its expected variation due to the annotation.
        \item The higher the gradient variation norm, the more significant is the impact of annotation on the transferability of representations.
    \end{enumerate}
    \item We follow \cite{ash2019deep} for increasing sampling diversity by promoting target samples for which SAGE span on diverse directions using the $\texttt{k-means++}$ initialization \cite{arthur2006k}. 
    \item We develop a novel training procedure to incorporate active target samples when learning domain invariant representations. It is split into two steps. The first step, called \textit{inductive step}, aims to update the classifier smoothly to reflect the annotation. A second step, called the \textit{transfer step}, leverages the classifier update for learning new class-level invariance.
    \item We conduct an empirical analysis on well-adopted benchmarks of UDA. It demonstrates that our approach improves the state-of-the-art significantly in Active Adversarial Domain Adaptation \cite{su2020active} on these datasets. All other things being equal, SAGE performs similarly or better than entropy-based uncertainty sampling or random sampling, making it a credible research direction for the design of new active DA algorithms.
\end{itemize}

The rest of the paper is organized as follows. First, we provide a brief overview of Domain Adversarial Learning for UDA. Importantly, we expose a soft-class conditioning adversarial loss, which reflects the transferability error of domain invariant representations \cite{bouvier2020robust}. Second, we motivate the use of Active Learning for enhancing their transferability. Third, we present the details of  Stochastic Adversarial Gradient Embedding. Finally, we conduct an empirical investigation on several benchmarks. 

\section{Background}
\label{background}

\paragraph{Notations.} Let us consider three random variables; $X$ the input data, $Z$ the representations and $Y$ the labels, defined on spaces $\mathcal X$, $\mathcal Z \subset \mathbb R^d$ where $d$ is the dimension of the representation, and $\mathcal Y$, respectively. We note realizations with lower cases, $x$, $z$ and $y$, respectively. Those random variables may be sampled from two and different distributions: the \textit{source} distribution $p_S(X,Z, Y)$ \ie data where the model is trained and the \textit{target} distribution $p_T(X,Z, Y)$ \ie data where the model is evaluated. Labels are one-hot encoded \ie $y \in [0,1]^{c}$ with $\sum_i y_i = 1$ where $c$ is the number of classes. We use the index notation $S$ and $T$ to differentiate source and target quantities. We define the hypothesis class $\mathcal H$ as a subset of functions from $\mathcal X$ to $\mathcal Y$ which is the composition of a representation class $\Phi$ (mappings from  $\mathcal X$ to $\mathcal Z$) and a classifier class $\mathcal F$ (mappings from $\mathcal Z$ to $\mathcal Y$) \ie $h:= f\varphi := f\circ \varphi \in \mathcal H$ where $f \in \mathcal F$ and $\varphi \in \Phi$. For $D \in \{S,T\}$ and an hypothesis $h \in \mathcal H$, we introduce the risk in domain $D$,  $\varepsilon_D(h) := \mathbb E_D[\ell(h(X),Y)] $ where $\ell$ is the $L^2$ loss $\ell(y,y') = ||y-y'||^2$ and $h(x)_i$ is the probability of $x$ to belong to class $i$. We note the source domain  data $(x_i
^S, y_i^S)_{1\leq i \leq n_S}$ and the target domain data $(x_i
^T)_{1\leq i \leq n_T}$.

\paragraph{Domain Adversarial Learning.} The seminal works from \cite{ganin2015unsupervised, ganin2016domain, long2015learning}, and their theoretical ground \cite{ben2007analysis, ben2010theory}, have led to a wide variety of methods based on domain invariant representations \cite{long2016unsupervised, long2017deep, long2018conditional, liu2019transferable, chen2019progressive, combes2020domain}. A representation $\varphi$ and a classifier $f$ are learned by achieving a trade-off between source classification error and domain invariance of representations by fooling a discriminator trained to separate the source from the target domain:
\begin{equation}
    \mathcal L(\varphi, f)  := \mathcal L_{\mathrm{c}}(\varphi, f) - \lambda\cdot \inf_{d \in \mathcal D} \mathcal L_{\mathrm{inv}}(\varphi, d)
\end{equation}
where $\mathcal L_{\mathrm{c}}(\varphi, f):= \mathbb E_{x,y \sim p_S}[ - y \cdot \log(f\varphi(x))]$ is the cross-entropy loss in the source domain, $\mathcal L_{\mathrm{inv}}(\varphi, d) := \mathbb E_{x \sim p_S}[\log (1 - d(\varphi(x)))] + \mathbb E_{x \sim p_T} [\log(d(\varphi(x))]$ is the adersarial loss and $\mathcal D$  is the set of discriminators \textit{i.e.} mapping from $\mathcal Z$ to $[0,1]$.  In practice, $\inf_{d \in \mathcal D}$ is approximated using a \textit{Gradient Reversal Layer} \cite{ganin2015unsupervised}. 

\paragraph{Transferability loss for class-level invariance.} Promoting class-level domain invariance improves transferability of representations \cite{long2018conditional}. Recently, \citeauthor{bouvier2020robust} introduce the \textit{transferability loss}, noted $\mathcal L_{\mathrm{tsf}}$, which performs class-conditioning in the adversarial loss by computing a scalar product between labels $y$ and a class-level discriminator $\mathsf d$ defined as a mapping from $\mathcal Z$ to $[0,1]^c$. Since labels are not available in the target domain at train time, predicted labels $\hat y := f\varphi(x)$ are used. This approach is referred to as \textit{soft-class conditioning}: 
\begin{equation}
    \mathcal L(\varphi, f) := \mathcal L_{\mathrm{c}}(\varphi, f) - \lambda \cdot \inf_{\mathsf d \in \mathsf D}\mathcal L_{\mathrm{tsf}}(\varphi, \hat y, \mathsf d)
    \label{overall_objective}
\end{equation}
where $\mathcal L_{\mathrm{tsf}}(\varphi, \hat y, \mathsf d) := \mathbb E_{(y, x) \sim p_S}  [ y \cdot \log( 1- \mathsf d(\varphi(x))) ] + \mathbb E_{(\hat y,x) \sim p_T}  [ \hat y \cdot \log(\mathsf d(\varphi(x))] $ is the transferability loss and $\mathsf D$ is the the set of class-level discriminators \textit{i.e.} mappings from $\mathcal Z$ to $[0,1]^c$. In this work, we explore the role of active annotation of a small subset of the target domain in order to improve the transferability of representations.

\section{Theoretical Analysis} 
\paragraph{Naive active classifier.} We consider $\mathcal A \subset \mathcal X$ a measurable subset of $\mathcal X$ with probability $b:= p_T(X \in \mathcal A)$. The subset $\mathcal A$ is given to an oracle which provides the ground-truth: $y \sim \mathrm{Oracle}(x)$ where $x \in \mathcal A$. In concrete terms, $b$ reflects our annotation budget. Given a hypothesis $h \in \mathcal H $, we introduce a naive classifier $h_{\mathcal A}$ that returns the predicted label $h(x)$ if $x$ is not annotated and the Oracle annotation if $x$ is in the subset $ \mathcal A$. Thus, \begin{equation}
    h_{\mathcal A}(x) = \mathrm{Oracle}(x) \mbox{ if } x \in \mathcal A, h(x) \mbox{ otherwise.}
    \label{eq:naive}
\end{equation}
To measure the quality of the active set $\mathcal A$, we introduce the notion of purity. In particular, we are interested in the amount of information coming from the Oracle. The purity is thus defined as $\pi:= p_T\left (h(X) \neq \mathrm{Oracle}(X)\right |X \in \mathcal A)$ and reflects our capacity to identify misclassified target samples.
With this notion, we observe the naive classifier improves the target error:
\begin{equation}
    \varepsilon_T\left (h_{\mathcal A}\right) \leq \varepsilon_T(h) - b \pi
    \label{error_naive_classifier}
\end{equation}
The higher the budget of annotation $b$ and the higher the purity $\pi$, the lower the target error of the naive classifier.

\paragraph{The naive classifier as a Transferability Inductive Bias.}
We now show how the target error of the naive classifier $\varepsilon_T\left (h_{\mathcal A}\right)$ is related to the source error of a classifier trained on source labelled data $\varepsilon_S(h)$, the annotation budget $b$, the purity of the annotated subset $\pi$, and the transferability of representations $\tau$. We build on the work of \citeauthor{bouvier2020robust} by interpreting the naive classifier as an inductive bias. More precisely, the naive classifier's target error is bounded as follows (see supplemental material for the proof):
\begin{equation}
    \varepsilon_T(h_{\mathcal A}) \leq \left (\frac{1}{b \pi} - 1\right) (\varepsilon_S(h)  + 8 \tau + \eta )
    \label{inductive_bound}
\end{equation}
where $\tau := \sup_{\mathsf f \in \mathsf F} \{ \mathbb E_{x\sim p_T}\left [h_{\mathcal A}(x) \cdot \mathsf f(\varphi(x))\right ] - \mathbb E_{x,y \sim p_S}[y\cdot \mathsf f(\varphi(x))] \}$ is the transferability error, $\mathsf F$ is the set of continuous functions from $\mathcal Z$ to $[-1, 1]^{C}$ and $\eta:=\inf_{\mathsf f \in \mathsf F}\varepsilon_T(\mathsf f\varphi)$. It is important to note that the transferability loss $\mathcal L_{\mathrm{tsf}}(\varphi, \hat y_{\mathcal A}, \mathrm d)$ is a domain adversarial proxy of the transferability error 
$\tau$ where $\hat y_{\mathcal A} = h_{\mathcal A}(x)$ \cite{bouvier2020robust}. Interestingly, target labels are only involved in $\pi$ and $\eta$ where the latter is an incompressible error that we assume to be small. The target error of the active classifier is a decreasing function of both the purity and the annotation budget and an increasing function of the transferability error.

\section{Proposed Method}
We first expose our motivations to embed target samples using the gradient of the transferability loss.  Indeed, this quantity allows to assess the impact of annotation on representations transferability. Second, we define the \textit{Stochastic Adversarial Gradient Embedding} (SAGE), an embedding where the norm quantifies this impact efficiently. Third, we  increase the diversity of sampling in this space, as described in \cite{ash2019deep}. Finally, we detail the procedure for the active learning of domain invariant representations.

\begin{figure*}
  \centering
  \subfigure[Positive orth. projection]{
    \label{fig:positive_proj}
    \includegraphics[width=.2\textwidth]{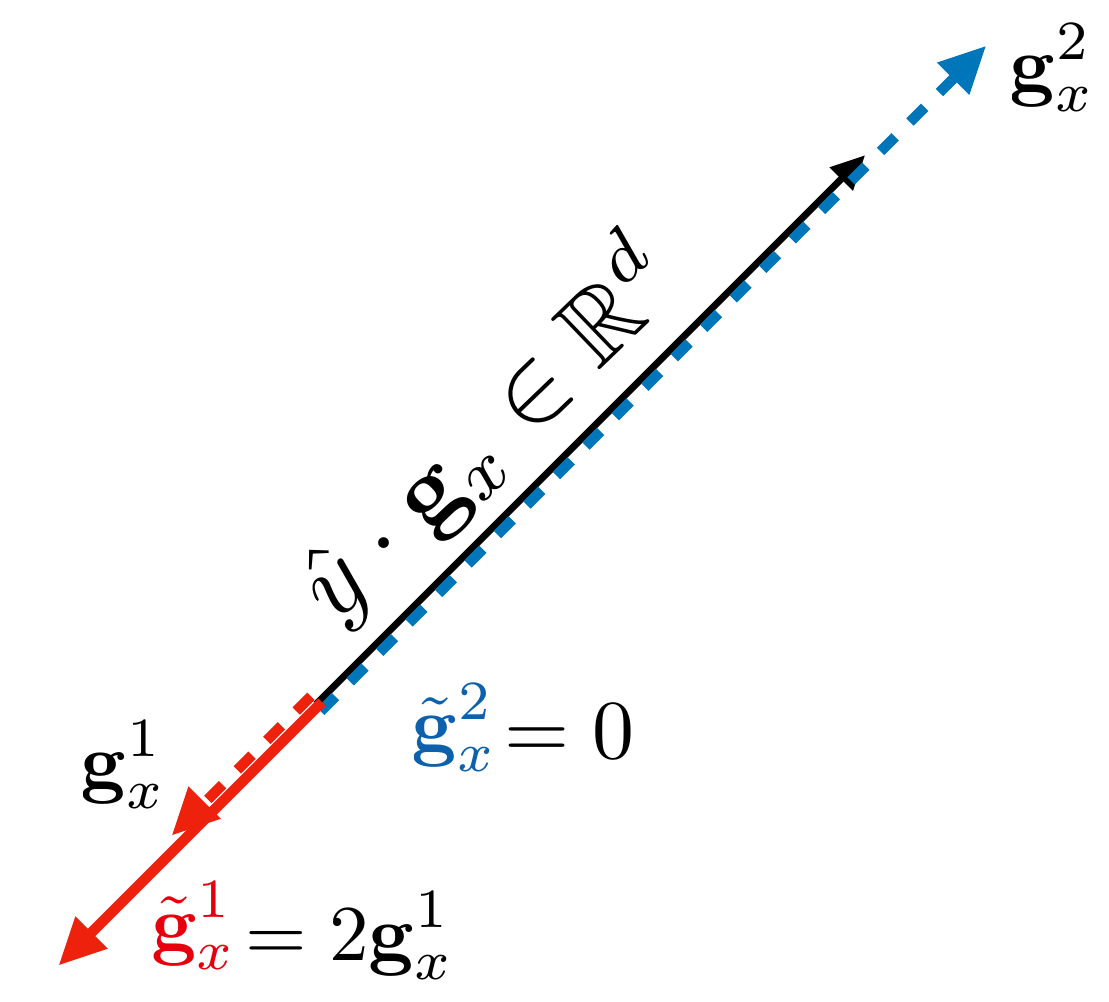}
  }
  \subfigure[SAGE]{
    \label{fig:SAGE}
    \includegraphics[width=.33\textwidth]{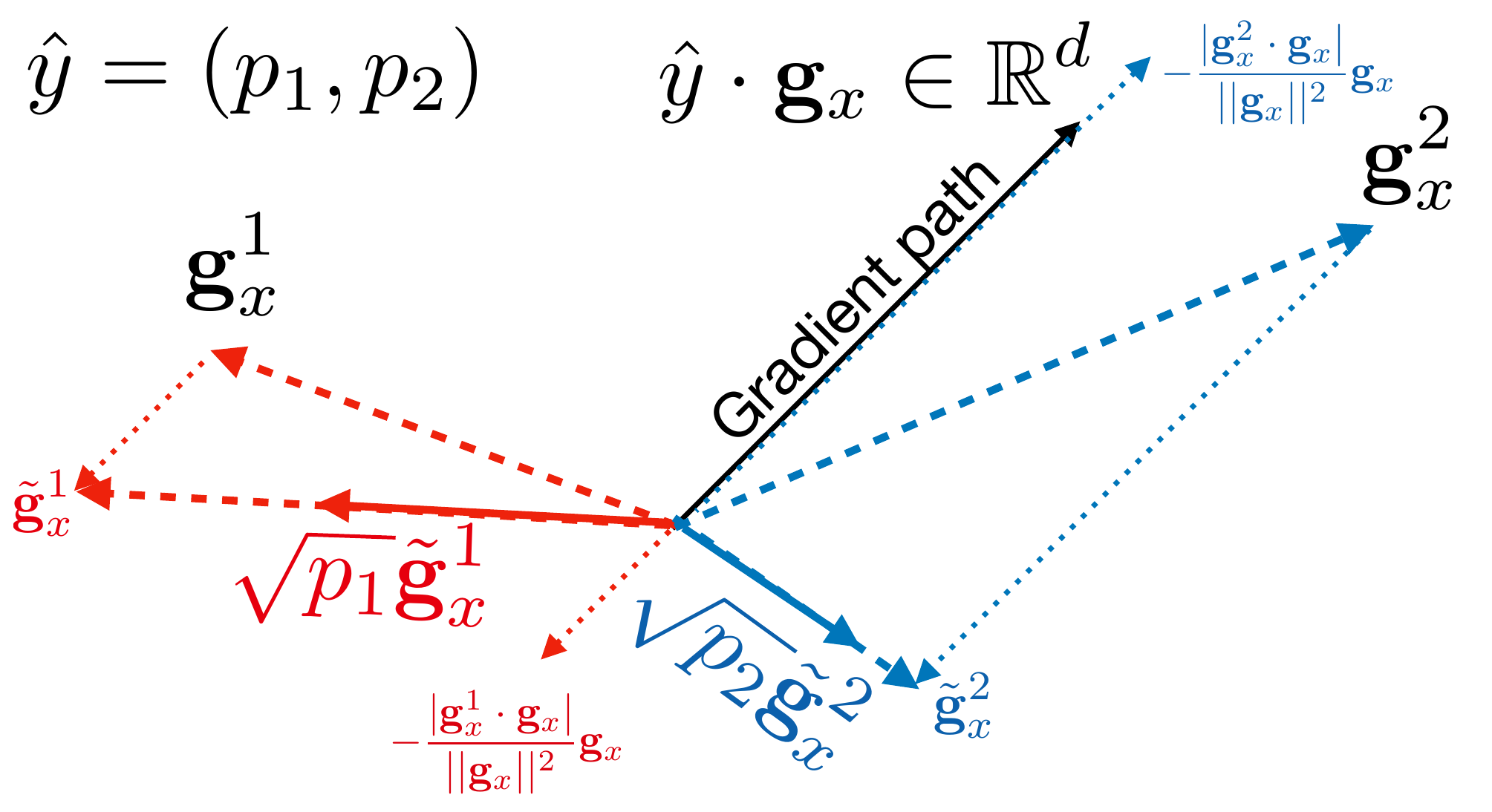}
  }
  \subfigure[Poor local minimum]{
    \label{fig:poor_min}
    \includegraphics[width=.25\textwidth]{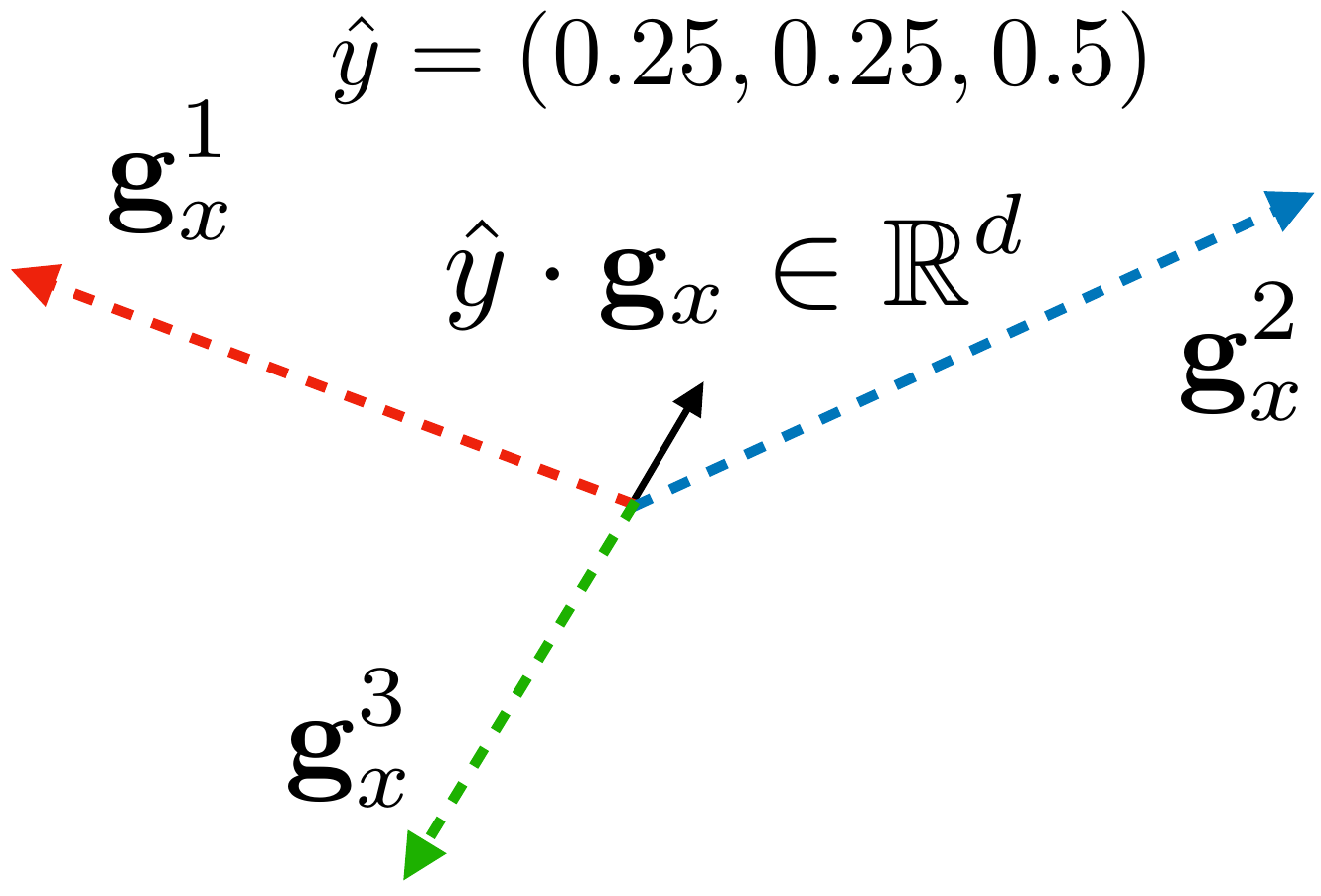}
  }

  \caption{(a) The \textit{positive} orthogonal projection cancels the gradient when the annotation agrees with the prediction. It doubles its norm when the annotation disagrees with the prediction.  (b) Visualisation of $\mathrm{SAGE}(x) = (\sqrt{p_1} \tilde{\mathbf g}_x^1, \sqrt{p_2} \tilde{\mathbf g}_x^2)$. Here $\tilde{\mathbf g}_x^2 \perp \mathbf g_x^2$ since $\hat y \cdot \mathbf g_x$ and $\mathbf g_x^2$ have a similar direction while $|\tilde{\mathbf g}_x^1 \cdot \mathbf g_x | \geq | \mathbf g_x
 ^1\cdot \mathbf g_x |$ since $\mathbf g_x^1$ as a component in the opposite direction of $\hat y \cdot \mathbf g_x$.  (c) Illustration of a case where the transferability loss is close to a local minimum ($\hat y \cdot \mathbf g_x \approx 0 $), but the stochastic gradients ($\mathbf g_x^y$ for $y \in \{1,2,3\}$) have a high norm. Here, the annotation chooses one of the gradients resulting in a strong update of the model.} 
  \label{fig:sensitivity}
\end{figure*}

\subsection{Motivations} As shown in the theoretical section, the budget $b$, the purity $\pi$ and the transferability of representations $\tau$ are levers to improve the naive classifier target error. The budget $b$ must be considered as a cost constraint and not as a parameter to be optimized. 
The purity of $\pi$ is not tractable since it involves labels in the target domain. 
Therefore, we focus our efforts on understanding the role of active annotation in improving transferability error $\tau$. Given a target sample $x \sim p_T(X)$ with representation $z:= \varphi(x) \in \mathbb R^d$, we expose the effect of annotating the sample $x$ on the gradient descent update of  \cref{overall_objective}. To conduct the analysis, we introduce the \textit{adversarial gradient} $\mathbf g_x$ of sample $x$ as the gradient of the discriminator loss with respect to the representation $z$:
\begin{equation}
    \mathbf g_x := - \frac{\partial \log(\mathsf d(z)) }{\partial z}\in \mathbb R^{c\times d}, ~~\mbox{where } \mathsf d(z) \in [0,1]^c
\end{equation}
Following the expression of the transferability loss $\mathcal L_{\mathrm{tsf}}$, the contribution of a sample $x$ to the gradient update (\cref{overall_objective}), before and after its annotation, is: 
\begin{equation*}
    \underbrace{\left \{ \theta \leftarrow \theta - \alpha \frac{\partial z}{\partial \theta} \cdot  \left( \hat y \cdot \mathbf g_x\right )\right \}}_{\mbox{\textit{Before annotation}}} \underset{}{\longrightarrow} \underbrace{\left \{ \theta \leftarrow \theta - \alpha \frac{\partial z}{\partial \theta} \cdot  \left( y \cdot \mathbf g_x\right )\right \} }_{\underset{y \sim \mathrm{Oracle}(x)}{\mbox{\textit{After annotation}}}}
    \label{oracle_gradient}
\end{equation*}
where $\partial z / \partial \theta$ is the jacobian of the representations with respect to the deep network parameters $\theta$ \ie $z:=\varphi_\theta(x)$, $\hat y :=  g\varphi_\theta(x)$ is the current label estimation and $\alpha$ is some scaling parameter. Before the annotation, the gradient vector can be written as a weighted sum of $\mathbf g_x$ \ie $\hat y \cdot \mathbf g_x \in\mathbb R^d$,  reflecting the class probability of $x$. Annotating the sample $x$ has the effect of setting, once and for all, a direction in $\mathbb R^c$ of the gradient ($y \cdot \mathbf g_x$). Based on this observation, we can measure the annotation procedure's ability to learn more transferable representation by its tendency to change the path of the gradient descent \ie how $y \cdot \mathbf g_x$ may differ with $\hat y \cdot \mathbf g_x$.

\subsection{Positive orthogonal projection} The adversarial gradient $\mathbf g_x:= \partial \log(\mathsf d(z))/\partial z\in \mathbb R^{c\times d}$ embodies the uncertainty on the true labels in the first dimension $\mathbb R^c$. In the rest of the paper, we now consider $\mathbf g_x$ as a stochastic vector of $\mathbb R^d$ with realizations lying in a discrete support $\mathcal G_x := \{\mathbf g_x^1, ..., \mathbf g_x^c\}$ where $\mathbf g_x^i = (\partial \log(\mathsf d(z))/\partial z)_i$. When provided the label through an oracle \ie $ y \sim \mathrm{Oracle}(x)$, we obtain $\mathbf g_x^y \in \mathcal G_x$, a realization of $\mathbf g_x$. Before annotation, the direction of the gradient is the mean of $\mathbf g_x$ where $\mathcal G_x$ is provided with a probability measure given by the classifier $h(x)$ \ie $p(\tilde{\mathbf g}_x = \tilde{\mathbf g}_x
^i) = h(x)_i$: 
\begin{equation}
    \mathbb E_h [\mathbf g_x] := \mathbb E_{y \sim h(x)} \left [\mathbf g_x^y \right ] \in \mathbb R^d
\end{equation}
Therefore, the tendency to modify the direction of the gradient is reflected by a high discrepancy between $\mathbb E_{h}[\mathbf g_x]$ and $\mathbf g_x^y$ for $y \sim \mathrm{Oracle}(x)$. To  quantify this discrepancy, we consider variations in both direction and magnitude: 
\begin{itemize}
    \item \textit{Direction:} A simple way to learn a new model by gradient descent is to find samples which modify the gradient's direction drastically.
    \item \textit{Magnitude:} The higher the norm of the gradient, the stronger the update of the model. 
\end{itemize}
To find a good trade-off between direction and magnitude, we remove the mean direction of the gradient $\mathbb E_h[\mathbf g_x]$ to $\mathbf g_x$ by computing a \textit{positive} orthogonal projection:
\begin{equation}
    \tilde {\mathbf g}_{x} := \mathbf g_x - \lambda \mathbb E_h[\mathbf g_x]
\end{equation}
where $\lambda := |\mathbf g_x \cdot \mathbb E_h[\mathbf g_x]| / ||\mathbb E_h[\mathbf g_x]||^2 $. Note that we use  $|\mathbf g_x \cdot \mathbb E_h[\mathbf g_x]|$, rather than $\mathbf g_x \cdot \mathbb E_h[\mathbf g_x]$ for the standard orthogonal projection, hence its name of \textit{positive} orthogonal projection. 

On the one hand, if the annotation provides a gradient with the same direction as the expected gradient \ie the annotation reinforces the prediction, $\tilde {\mathbf g}_{x}$ is null. On the other hand, if the annotation provides a gradient with an opposite direction to the expected gradient \ie the annotation contradicts the prediction, the norm of $\tilde {\mathbf g}_{x}$ increases. Therefore, target samples $x$ for which we expect the highest impact on the transferability, are those with the highest norm of $\tilde{\mathbf g}_x$. An illustration is provided in \cref{fig:positive_proj}. Since $\tilde{\mathbf g}_x$ is random, we need additional tools to define a norm operator properly on it,  leading to our core contribution.

\subsection{Stochastic Adversarial Gradient Embedding}
A simple way to define the norm of $\tilde{\mathbf g}_x$ would be to consider $|| \tilde{\mathbf g}_x ||_h: =  ( \mathbb E_{y \sim h(x)} \left [|| \tilde{\mathbf g}_x^y ||^2\right])^{1/2}$ \ie the expected norm of $\tilde{\mathbf g}_x$. This leads to a "distance" defined as follows $(\mathbb E_{y_1\sim h(x_1), y_2\sim h(x_2) } \left [ ||\mathbf g_{x_1}^{y_1} - \mathbf g_{x_2}^{y_2} ||^2 \right] )^{1/2}$. However, it is straightforward to observe such a "distance" between $\tilde{\mathbf g}_x$ and itself is not null if $h(x)$ is not a one-hot vector; not making it, in fact, a proper distance.  

To address this issue, we suggest to embed, through a mapping named \textit{Stochastic Adversarial Gradient Embedding} ($\mathrm{SAGE}$), the coupling $(h, \tilde{\mathbf g})$ in a vectorial space by considering the following tensorial product:
\begin{equation}
    \mathrm{SAGE}(x) := (\sqrt{h(x)_1} \tilde{\mathbf g}_x^1, ..., \sqrt{h(x)_c} \tilde{\mathbf g}_x^c) \in \mathbb R^{c \times d}
\end{equation}
Importantly, the choice of using $\sqrt h$ is motivated by the observation that $||\mathrm{SAGE}(x) || = || \tilde{\mathbf g}_x||_h$ leading to a proper distance between $x_1$ and $x_2$: 
\begin{equation}
    \Delta_h(x_1, x_2) := || \mathrm{SAGE}(x_1) - \mathrm{SAGE}(x_2)||
\end{equation}
Crucially, both the norm and the distance computed on SAGE do not involve the target labels, making it relevant in the UDA setting where target labels are unknown. An illustration of SAGE is provided in \cref{fig:SAGE}.

\subsection{Increasing Diversity of SAGE} As aforementioned, the higher the norm of $||\mathrm{SAGE}(x)||$, the greater the expected impact of annotating sample $x$ on the transferability of representations. A naive strategy of annotation would be simply to rank target samples by their SAGE norm ($||\mathrm{SAGE}(x)||$). However, this strategy ignores the crucial problem of \textit{diversity} when running active annotation \cite{settles2009active}. Embedding target samples offers the opportunity to increase sampling diversity by selecting samples with high expected norms and various directions \cite{ash2019deep}. The $\texttt{k-means++}$ initialization is known to select diverse and high norm vectors. Roughly speaking, the algorithm starts by selecting the vectors $v$ with the highest norm and the second, $v'$, such that $v' -v$ has the highest norm, and so on. The procedure is detailed in Algorithm 1.

\begin{algorithm}[ht!]
\caption{\texttt{diverse\_SAGE}$((x_i^T)_{1 \leq i \leq n_T}, f,  \varphi,\mathsf d, b)$}
\textbf{Input}: Target samples $(x_i^T)_{1 \leq i \leq n_T}$, classifier $f$, representation $\varphi$,  discriminator $\mathsf d$,  budget $b$:
\begin{algorithmic}[1] 
\STATE $\mathcal A \leftarrow \mathtt{[}\arg \max_{1 \leq i \leq n_T} ||\tilde{\mathbf g}_{x_i^T}||_h\mathtt{]}$
\WHILE{$\mathtt{len}(\mathcal A) < b$}
\STATE $\mathcal A\mathtt{.append} \left ( \displaystyle{ \arg \max_{1 \leq i \leq n_t} \min_{ a \in \mathcal A} \Delta_{f\varphi}(x_i^T, x_{a}^T) } \right)$
\ENDWHILE
\STATE \textbf{Return} $\mathcal A$
\end{algorithmic}
\end{algorithm}

\subsection{Training procedure}
The training procedure is described in Algorithm 2. First, we train the model by UDA \cite{bouvier2020robust}. Second, for a given number of iterations, we select by SAGE $b$ samples to send to the Oracle. Then, we perform two steps: an \textit{inductive step} and a \textit{transfer step}. The former updates the current classifier, for learning an \textit{active classifier}, by incorporating the knowledge provided by annotated samples. The latter updates the representations for achieving class-level invariance where the predictions of the active classifier are used in the target domain. This procedure is repeated for the $r$ annotation rounds. In the following, we note $h_a \in \mathcal H$ an \textit{active classifier} \ie a classifier that takes into account annotations provided by the Oracle (\eg the naive classifier).

\paragraph{Transfer step.} Based on our theoretical analysis, the representations transferability is improved when $h_a$ is introduced into the transferability loss:
\begin{align}
    \notag \mathcal L_{\mathrm{tsf}}(\varphi, h_a, \mathsf d) := & ~ \mathbb E_{(y, x) \sim p_S}  [ y \cdot \log( 1- \mathsf d(\varphi(x))) ] \\ 
    & ~ + \mathbb E_{x\sim p_T}  [ h_a(x) \cdot \log(\mathsf d(\varphi(x))]
\end{align}
The active classifier is involved in the target domain to compute the transferability loss. Therefore, we introduce the \textit{Transfer step} which consists in the following stochastic gradient descent update:
\begin{align}
    \notag \mbox{\texttt{transfer}}(f,\varphi, h_a) := & (f,g) -  \alpha \nabla_{(f,g)} \{ \mathcal L_{\mathrm{c}}(\varphi, f)  \\
    & -  \lambda \cdot \inf_{\mathsf d \in \mathsf D}\mathcal L_{\mathrm{tsf}}(\varphi, h_a, \mathsf d) \} 
\end{align}
where $\inf_{\mathsf d \in \mathsf D}$ is in practice a gradient reversal layer \cite{ganin2015unsupervised}, $\alpha$ is scaling parameter, $\lambda$ varies smoothly from $0$ to $1$ during training as described in \cite{long2018conditional} and losses are computed on batches of samples. 

\begin{figure}[ht!]
    \centering
    \includegraphics[scale=0.35]{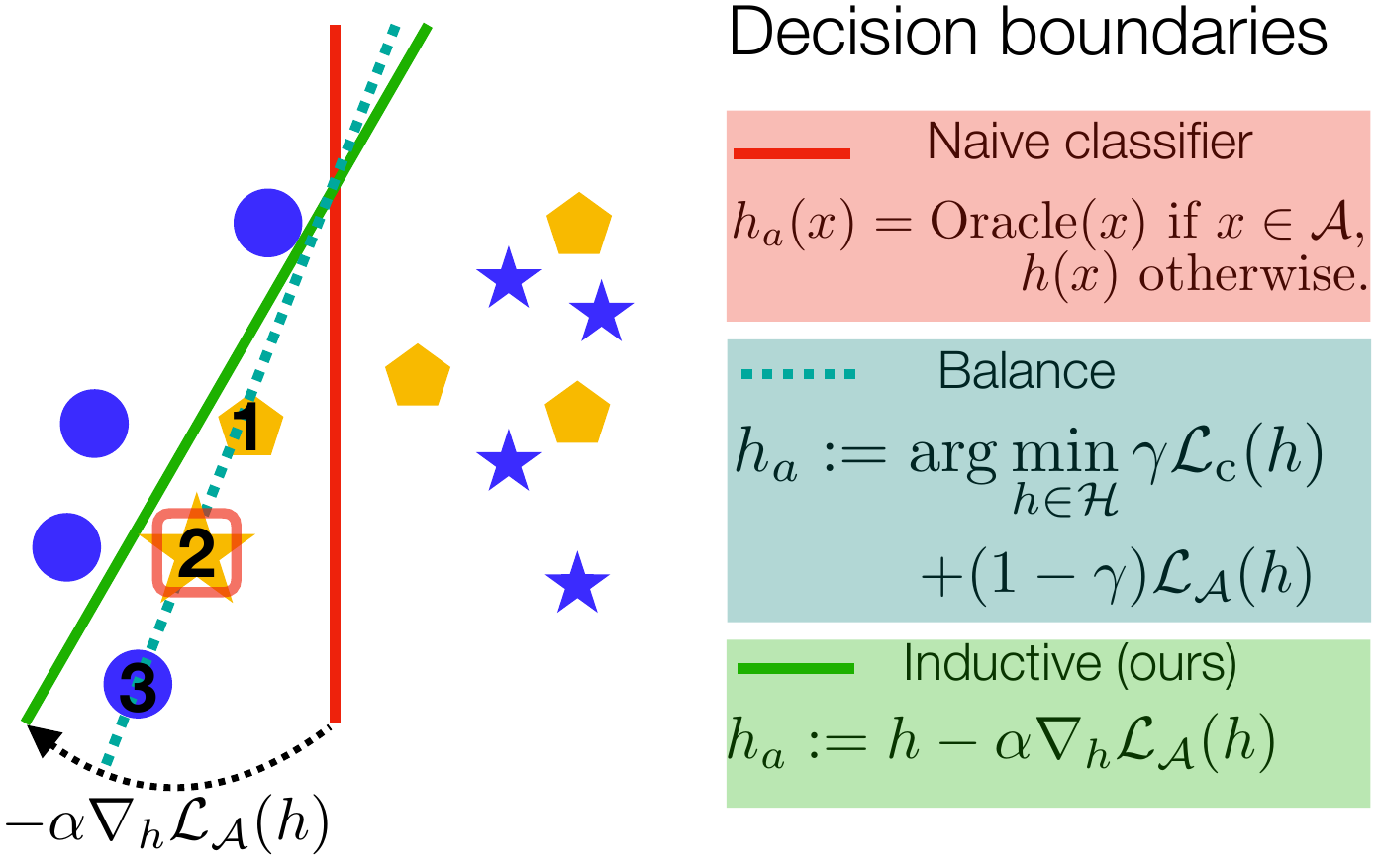}
    \caption{Illustration of the decision boundary update when varying the inductive design of the active classifier. Sample \textbf{2} is sent to an Oracle providing annotation $\star$ when the current classifier $h$ predicts $\bullet$. Based on this annotation, one can assume strongly that \textbf{1} is a $\star$. Here, the naive classifier predicts $\bullet$ for $\textbf{1}$. Both \textbf{1} and \textbf{2} are close to the balance classifier's decision boundary due to a $\bullet$ source sample \textbf{3} resulting in an uncertain predictions, thus a poor class-conditioning in the transferability loss. Our inductive step allows to obtain confident predictions for both \textbf{1} and \textbf{2}. The active classifier misclassifies \textbf{3} enforcing the model to learn a better-suited representation for \textbf{3}, improving the representation alignment.}
    \label{fig:inductive_decision_boundary}
\end{figure}

\begin{figure*}
  \centering
  \subfigure[A$\to$W]{
    \label{fig:AW}
    \includegraphics[width=.148\textwidth]{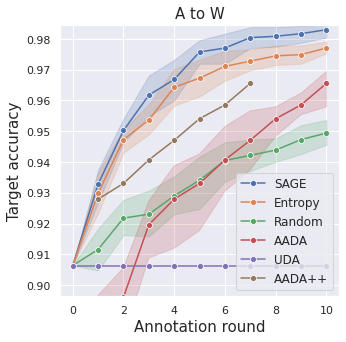}
  }
  \subfigure[W$\to$A]{
    \label{fig:WA}
    \includegraphics[width=.148\textwidth]{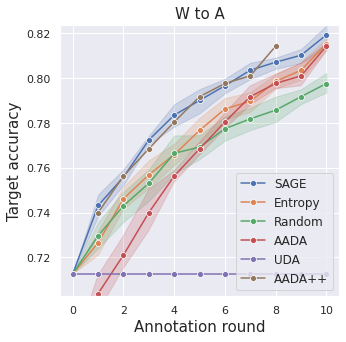}
  }
  \centering
  \subfigure[A$\to$D]{
    \label{fig:AD}
    \includegraphics[width=.148\textwidth]{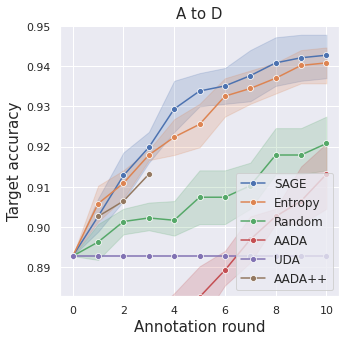}
  }
  \subfigure[D$\to$A]{
    \label{fig:DA}
    \includegraphics[width=.148\textwidth]{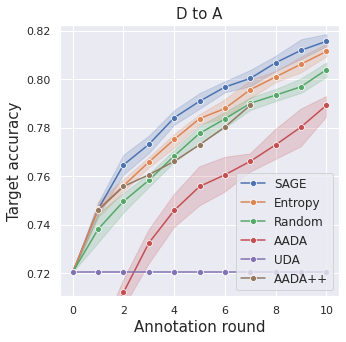}
  }
\subfigure[ VisDA($b=10$)]{
    \label{fig:visda_10}
    \includegraphics[width=.148\textwidth]{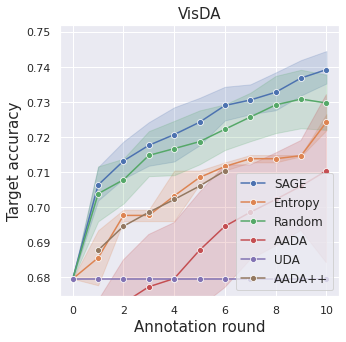}
  }
 \subfigure[VisDA($b=100$)]{
    \label{fig:visda_100}
    \includegraphics[width=.148\textwidth]{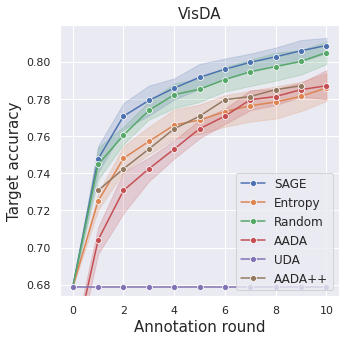}
  }

  \caption{SAGE outperforms the baselines for the six tasks. Annotation of target samples improves the transferability of domain invariant representations drastically. If \textit{uncertainty-based} selection (\textbf{Entropy}) performs better than \textit{diversity-based} selection (\textbf{Random}) for the Office31 dataset, the opposite is observed for the VisDA dataset. For both datasets, SAGE performs better than both uncertainty and diversity-based selections, demonstrating its capacity to take the best of the two worlds.} 
  \label{fig:results}

\end{figure*}

\begin{algorithm}
\caption{Training procedure}
\textbf{Input}: Target samples $(x_i^T)_{1 \leq i \leq n_T}$, annotation budget $b$, annotation rounds $r$, iterations $n_{\mathrm{it}}$:
\begin{algorithmic}[1] 
\STATE $f, \varphi \leftarrow$ UDA pretraining, $\mathcal A \leftarrow \mathtt{[} ~\mathtt{]} $. 
\FOR{$r$ rounds of annotations}
\STATE $\mathcal A\mathtt{.append(}  \mbox{\texttt{diverse\_SAGE}}((x_i^T)_{1 \leq i \leq n_T}, f,\varphi, \mathsf d, b) \mathtt{)}$ \# \textit{Annotate target samples selected by diverse SAGE.}
\FOR{ $n_{\mathrm{it}}$ iterations}
\STATE $h_a \leftarrow \mbox{\texttt{inductive}}(f\varphi, \mathcal A)$ \# \textit{Inductive step: annotations are incorporated in the active classifier.}
\STATE $(f, \varphi) \leftarrow  \mbox{\texttt{transfer}}(f,\varphi, h_a)$ \# \textit{Transfer step: representation and classifier update for aligning with the feedback of the active classifier.}
\ENDFOR
\ENDFOR
\label{alg:training_procedure}
\end{algorithmic}
\end{algorithm}

\paragraph{Inductive step.} We now focus our attention on the design of the active classifier $h_a$, referred to as the \textit{Inductive step}, as described in \cite{bouvier2020robust}. Our theoretical analysis from \cref{inductive_bound} holds for the naive classifier (it outputs the oracle annotation if the target sample is annotated and the current prediction otherwise). However, given two samples $x_1$ and $x_2$ close in the representation space \ie $z_1 \approx z_2$, such that $x_1$ is annotated, $y_1 \sim \mathrm{Oracle}(x_1)$, one can assume that the probability of observing $y_1 = \mathrm{Oracle}(y_2)$ is high. However, the design of the naive classifier does not reflect this inductive bias since $h_{\mathcal A}(x_2) = h(x_2)$ (see \cref{fig:inductive_decision_boundary}). Therefore, we suggest to train an active classifier based on the annotation provided by the Oracle to spread the information in the representation space neighborhood of $x_1$. We propose to use a simple loss that incorporates both the error in the source domain and the error in target annotated samples as follows:
\begin{equation}
    h_a := \arg \min_{h \in \mathcal H} \gamma \mathcal L_{\mathrm{c}}(h)  + (1-\gamma){\mathcal L}_{\mathcal A}(h)
    \label{eq:balance}
\end{equation}
where ${\mathcal L}_{\mathcal A}(h) = \mathbb E_{x \in \mathcal A}[ - \mathrm{Oracle}(x) \cdot \log(h(x))]$ and $\gamma \in (0,1)$ is a trade-off parameter. On the one hand, when the annotated samples have high importance to learn $h_a$ \ie $\gamma$ tends to 0, we are exposed to a high risk of high variance of such classifier since $|\mathcal A| \ll n_S$.
On the other hand, when the source samples have a high importance to learn $h_a$ \ie $\gamma$ tends to 1, knowledge provided by the annotation is poorly learned by the active classifier. Therefore, calibrating properly $\gamma$ is a challenging problem. We overcome this issue by smoothly updating the classifier $h$ as follows:
\begin{equation}
 \mbox{\texttt{inductive}}(h, \mathcal A) := h - \alpha \nabla_h {\mathcal L}_{\mathcal A}(h)
 \label{eq:fine_tune}
\end{equation}
The design of \texttt{inductive} aims not to forget knowledge acquired in the source domain while integrating the knowledge provided by the annotation of the target samples.

\section{Experiments}

\paragraph{Datasets.} We evaluate our approach on \textbf{Office-31} \cite{saenko2010adapting} and \textbf{VisDA-2017} \cite{peng2017visda}. Office-31 contains 4,652 images classified in 31 categories across three domains: Amazon (\textbf{A}), Webcam (\textbf{W}), and DSLR (\textbf{D}). We explore tasks $\textbf{A} \to \textbf{W}$, $\textbf{W} \to \textbf{A}$, $\textbf{A} \to \textbf{D}$ and $\textbf{D} \to \textbf{A}$. We do not report results for tasks $\textbf{D} \to \textbf{W}$ and $\textbf{W} \to \textbf{D}$ since these tasks have already nearly perfect results in UDA \cite{long2018conditional}. 
For VisDA, we explore \textbf{Synthetic}: 3D models with different lightning conditions and different angles; \textbf{Real}: real-world images. We explore the \textbf{Synthetic} $\rightarrow$ \textbf{Real} task. The standard protocol in UDA uses the same target samples during train and test phases. In the context of active learning, this induces an undesirable effect where sample annotation mechanically increases  the accuracy; at train time, the model has access to input and label of annotated samples which are also present at test time. We suggest instead to split the target domain into a \textit{train target domain} (samples used for adaptation and pool of data used for annotation) and \textit{test target domain} (samples used for evaluating the model) with a ratio of $1/2$. Therefore, samples from the test target domain have never been seen at train time. 

\paragraph{Setup.} For classification, we use the same hyperparameters than \cite{long2018conditional} and adopt ResNet-50 \cite{he2016deep} as a base network pre-trained on ImageNet dataset \cite{deng2009imagenet}. Our code is based on official implementations of \citeauthor{bouvier2020robust} derived from \cite{long2018conditional}. For all experiments, we have fixed $r=10$ rounds of annotations. For Office31, we have fixed a budget of $b=2\%$ of annotation of the train target domain \ie $20\%$ of the train target domain is annotated at the end of the 10 rounds. For VisDA, we have explored two budgets: $b=10$ or $100$ samples. We report average results obtained with 6 random experiments. We perform 10k iterations of SGD for the UDA pre-training while, between each annotation round, we perform 5k iterations of SGD. One experiment lasts about $\sim$12 hours on a single NVIDIA V100 GPU with 32GB memory. The full implementation is provided in the supplemental material.

\paragraph{Baselines.} \textbf{AADA} \cite{su2020active} is the closest algorithm to SAGE and the most interesting to compare. AADA learns domain invariant representations by fooling a domain discriminator $d$ trained to output 1 for source data and 0 for target data \cite{ganin2015unsupervised} and scores target samples $x$; $s(x) := H(\hat y) w(z)$ where $H(\hat y)$ is the entropy of predictions $\hat y$ and $w(z) = (1-d(z))/d(z)$. $H(\hat y)$ brings information about uncertainty while $w(z)$ brings diversity to the score. We have reproduced the implementation of AADA. SAGE starts active learning with a serious advantage to AADA as the pre-training procedures differ significantly. In order to get a fairer comparison with SAGE, we have therefore chosen to report a modified version of AADA that we call AADA++. AADA++ is free of charge as long as the performance is below the UDA baseline \cite{bouvier2020robust} (\textbf{UDA}). In our view, this should essentially eliminate AADA's structural disadvantage. 
In practical terms, we translate AADA to the left of the graph $\mathrm{(Annotation~round , Target~accuracy)}$ until accuracy at request 0 is higher than UDA, explaining why annotation rounds of AADA++ do not reach 10. We also report \textbf{Entropy} (\textit{uncertainty-based} sampling which selects samples with the highest entropy) and \textbf{Random} (\textit{diversity-based} sampling which selects samples randomly). For both Entropy and Random baselines, the training procedure of Algorithm 2 is followed, except for line \texttt{3} where SAGE is replaced by entropy sampling \cite{wang2014new} or  random sampling.

\paragraph{Results.} We report the results of experiments in \cref{fig:results}. Approximately 110 days of GPU time are necessary for reproducing the results. First, active annotation brings substantial improvements to UDA for both datasets. This validates the effort and the focus that should be put on active domain adaptation in our opinion. For the six tasks, SAGE outperforms the baselines (AADA in particular) with a comfortable margin. More precisely, the saturation regime for AADA for A$\to$W, D$\to$A and VisDA tasks is significantly below than the saturation regime of SAGE. Even when provided with free annotation rounds, AADA++ is less accurate than SAGE, except for task W$\to$A, demonstrating the significant improvement made by SAGE. In addition, we observe that AADA is unstable when only few data are annotated (VisDA($b=10$)), while SAGE remains robust in this regime. Finally, uncertainty-based sampling (Entropy) performed better than diverse-based sampling (Random) for the Office31 dataset, while the contrary is observed for the VisDA dataset. Interestingly, SAGE performs better than both Entropy and Random, showing that SAGE takes the best of both worlds (uncertainty \textit{vs} diversity). 

\paragraph{Ablation study.} We conduct an ablation study to compare the \textit{inductive step} described in \cref{eq:fine_tune} with a step based on \cref{eq:balance} (\textbf{Balance} with $\gamma=0.5$) or based on \cref{eq:naive} (\textbf{Naive}). We report results for tasks A$\to$W, W$\to$A and both tasks of VisDA($b=100$). We observe that both \textbf{Balance} and \textbf{Inductive} improve significantly performances compared to the \textbf{Naive} classifier, demonstrating the importance of the inductive step. For both tasks A$\to$W and W$\to$A, \textbf{Inductive} is slightly better than \textbf{Balance}, confirming our belief that smooth updating the classifier (as described in \cref{eq:fine_tune}) improves performances. Interestingly, \textbf{Inductive} performed better than Balance when few data are annotated (\eg VisDA with $b=10$, VisDA with $b=100$ for rounds lower than 3), while the contrary was observed when more data is annotated. This findings tend to show that \textbf{Inductive} is more adapted in the low annotation regime. All things considered, the difference between \textbf{Balance} and \textbf{Inductive} remains small compared to the improvement provided by SAGE compared to AADA. Nevertheless, the inductive step remains an important step in SAGE and deserves a deeper understanding. More ablation is provided in the Appendix.

\begin{figure}[ht!]
\centering
  \subfigure[A$\to$W]{
    \label{fig:AW_inductive}
    \includegraphics[width=.16\textwidth]{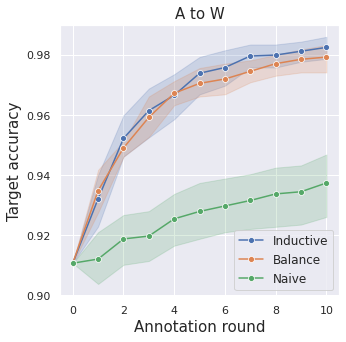}
  }
  \subfigure[W$\to$A]{
    \label{fig:WA_inductive}
    \includegraphics[width=.16\textwidth]{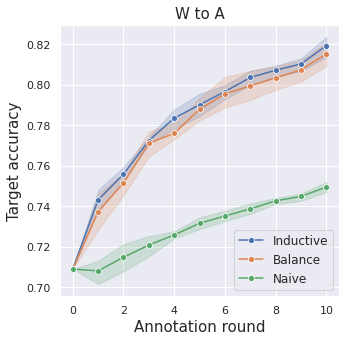}
  }
  \subfigure[VisDA($b=10$)]{
    \label{fig:visda_10_inductive}
    \includegraphics[width=.16\textwidth]{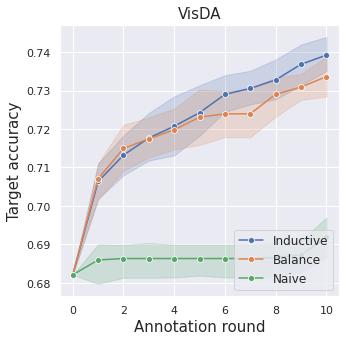}
  }
  \subfigure[VisDA($b=100$)]{
    \label{fig:VisDA_inductive}
    \includegraphics[width=.16\textwidth]{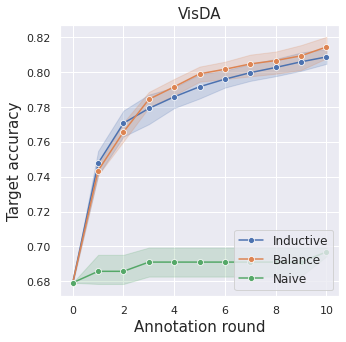}
  }

  \label{fig:ablation}
  \caption{Both \textbf{Inductive} and \textbf{Balance} improve the active learning compared with the \textbf{Naive} classifier. \textbf{Inductive} seems to provide better guarantees compared with the \textbf{Balance} classifier when only few data are annotated. This remark opens the way to interesting future work in ADA.} 
\end{figure}

\begin{figure}[ht!]
\centering
  \subfigure[A$\to$W]{
    \label{fig:AW_diverse}
    \includegraphics[width=.16\textwidth]{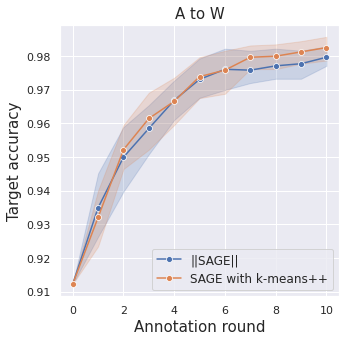}
  }
  \subfigure[W$\to$A]{
    \label{fig:WA_diverse}
    \includegraphics[width=.16\textwidth]{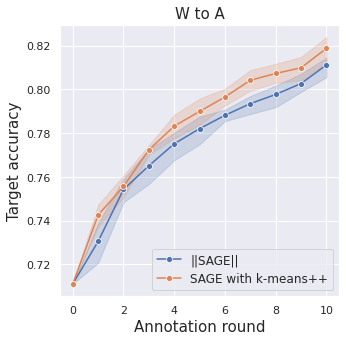}
  }
  \subfigure[VisDA($b=10$)]{
    \label{fig:VisDA_10_diverse}
    \includegraphics[width=.16\textwidth]{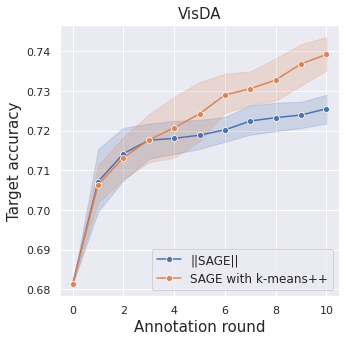}
  }
  \subfigure[VisDA($b=100$)]{
    \label{fig:VisDA_100_diverse}
    \includegraphics[width=.16\textwidth]{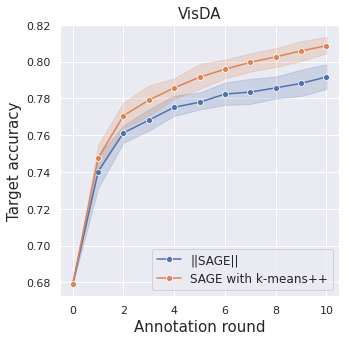}
  }
  \caption{For the four tasks selected for the ablation study, the \texttt{k-means++} initialization brings substantial improvments especially for tasks W$\to$A and VisDA for $b=10$ and $100$.} 
  \label{fig:ablation_diverse}
\end{figure}

We demonstrate the effectiveness of \texttt{k-means++} \cite{arthur2006k} in SAGE. Additionally to report SAGE (SAGE with \texttt{k-means}++), we also reports results of Active Domain Adaptation when target samples are selected with respect to their SAGE norm ($||\mathrm{SAGE}||$), thus not taking in account directions of gradients. Results are presented in \cref{fig:ablation_diverse} for tasks A$\to$W, W$\to$A and both tasks of VisDA($b=10$ and $100$).

\section{Related works}
\paragraph{Transferability of Invariant Representations.} Recent works warn that domain invariance may deteriorate transferability of invariant representations \cite{johansson2019support, zhao2019learning}. Prior works enhance their transferability with multi-linear conditioning of representations with predictions \cite{long2018conditional}, by introducing weights \cite{cao2018partial, bouvier2019hidden, you2019universal, zhang2018importance, combes2020domain}, by penalizing high singular value of representations batch \cite{chen2019transferability}, by hallucinating consistent target samples for bridging the domain gap \cite{liu2019transferable}  or by enforcing target consistency through augmentations which conserve the semantic of the input \cite{ouali2020target}. 

 \paragraph{Active Learning.} There is an extensive literature on Active Learning \cite{settles2009active} that can be divided into two schools; \textit{uncertainty} and \textit{diversity}. The first aims to annotate samples for which the model has uncertain prediction \eg samples are selected according to their entropy \cite{wang2014new} or prediction margin \cite{roth2006margin}, with some theoretical guarantees \cite{hanneke2014theory, balcan2009agnostic}. The second focuses on annotating a representative sample of the data distribution \eg the Core-Set approach \cite{sener2017active} selects samples that geometrically cover the distribution. Several approaches also propose a trade-off between uncertainty and diversity, \eg \cite{hsu2015active} that is formulated as a bandit problem. Recently, \citeauthor{ash2019deep} introduced BADGE, a gradient embedding, which, like SAGE, takes the best of uncertainty and diversity. Our work is inspired by BADGE and adapts the core ideas in the context of learning domain invariant representations.

 \paragraph{Active Domain Adaptation.} Despite its great practical interest, only a few previous works address the problem of \textit{Active Domain Adaptation}. \cite{chattopadhyay2013joint} annotates target samples by importance sampling while ALDA \cite{rai2010domain, saha2011active} annotates samples with high discrepancy with source samples based on the prediction of a domain discriminator. However, those strategies do not fit modern adaptation with deep nets. To our knowledge, AADA \cite{su2020active} is the only prior work that learns actively domain invariant representations and achieves the state-of-the-art for Active Domain Adaptation. Thus, AADA is the most relevant work to compare with SAGE.

\section{Conclusion}
We have introduced SAGE, an efficient method for active adversarial domain adaptation. SAGE is an embedding suitable for identifying target samples that are likely to improve representations' transferability when annotated. It relies on three core components; a stochastic embedding of the gradient of the transferability loss, a \texttt{k-means++} initialization which guarantees that each annotation round annotates a diverse set of target samples, and a two-step learning procedure that incorporates efficiently active target samples when learning invariant representations. Through various experiments, we have demonstrated the effectiveness of SAGE for improving the transferability of representations and its capacity to take the best of uncertainty and diversity sampling.

\section*{Acknowledgements}
Victor Bouvier is funded by Sidetrade and ANRT (France) through a CIFRE collaboration with CentraleSupélec. This work was performed using HPC resources from the “Mésocentre” computing center of CentraleSupélec and École Normale Supérieure Paris-Saclay supported by CNRS and Région Île-de-France (\url{http://mesocentre.centralesupelec.fr/}). We thank the contributors of \cite{walt2011numpy} and \cite{sklearn_api} that were used for \texttt{k-means++} initialization, \cite{Hunter:2007} for figures rendering and \cite{paszke2019pytorch} that was used as Deep Learning framework.

\bibliography{Bibliography-File}

\begin{thebibliography}{52}
\providecommand{\natexlab}[1]{#1}
\providecommand{\url}[1]{\texttt{#1}}
\providecommand{\urlprefix}{URL }
\expandafter\ifx\csname urlstyle\endcsname\relax
  \providecommand{\doi}[1]{doi:\discretionary{}{}{}#1}\else
  \providecommand{\doi}{doi:\discretionary{}{}{}\begingroup
  \urlstyle{rm}\Url}\fi

\bibitem[{Amodei et~al.(2016)Amodei, Olah, Steinhardt, Christiano, Schulman,
  and Man{\'e}}]{amodei2016concrete}
Amodei, D.; Olah, C.; Steinhardt, J.; Christiano, P.; Schulman, J.; and
  Man{\'e}, D. 2016.
\newblock Concrete problems in AI safety.
\newblock \emph{arXiv preprint arXiv:1606.06565} .

\bibitem[{Arjovsky et~al.(2019)Arjovsky, Bottou, Gulrajani, and
  Lopez-Paz}]{arjovsky2019invariant}
Arjovsky, M.; Bottou, L.; Gulrajani, I.; and Lopez-Paz, D. 2019.
\newblock Invariant Risk Minimization.
\newblock \emph{arXiv preprint arXiv:1907.02893} .

\bibitem[{Arthur and Vassilvitskii(2006)}]{arthur2006k}
Arthur, D.; and Vassilvitskii, S. 2006.
\newblock k-means++: The advantages of careful seeding.
\newblock Technical report, Stanford.

\bibitem[{Ash et~al.(2019)Ash, Zhang, Krishnamurthy, Langford, and
  Agarwal}]{ash2019deep}
Ash, J.~T.; Zhang, C.; Krishnamurthy, A.; Langford, J.; and Agarwal, A. 2019.
\newblock Deep batch active learning by diverse, uncertain gradient lower
  bounds.
\newblock \emph{arXiv preprint arXiv:1906.03671} .

\bibitem[{Balcan, Beygelzimer, and Langford(2009)}]{balcan2009agnostic}
Balcan, M.-F.; Beygelzimer, A.; and Langford, J. 2009.
\newblock Agnostic active learning.
\newblock \emph{Journal of Computer and System Sciences} 75(1): 78--89.

\bibitem[{Beery, Van~Horn, and Perona(2018)}]{beery2018recognition}
Beery, S.; Van~Horn, G.; and Perona, P. 2018.
\newblock Recognition in terra incognita.
\newblock In \emph{Proceedings of the European Conference on Computer Vision
  (ECCV)}, 456--473.

\bibitem[{Ben-David et~al.(2010)Ben-David, Blitzer, Crammer, Kulesza, Pereira,
  and Vaughan}]{ben2010theory}
Ben-David, S.; Blitzer, J.; Crammer, K.; Kulesza, A.; Pereira, F.; and Vaughan,
  J.~W. 2010.
\newblock A theory of learning from different domains.
\newblock \emph{Machine learning} 79(1-2): 151--175.

\bibitem[{Ben-David et~al.(2007)Ben-David, Blitzer, Crammer, and
  Pereira}]{ben2007analysis}
Ben-David, S.; Blitzer, J.; Crammer, K.; and Pereira, F. 2007.
\newblock Analysis of representations for domain adaptation.
\newblock In \emph{Advances in neural information processing systems},
  137--144.

\bibitem[{Bouvier et~al.(2020)Bouvier, Very, Chastagnol, Tami, and
  Hudelot}]{bouvier2020robust}
Bouvier, V.; Very, P.; Chastagnol, C.; Tami, M.; and Hudelot, C. 2020.
\newblock Robust Domain Adaptation: Representations, Weights and Inductive
  Bias.
\newblock \emph{arXiv preprint arXiv:2006.13629} .

\bibitem[{Bouvier et~al.(2019)Bouvier, Very, Hudelot, and
  Chastagnol}]{bouvier2019hidden}
Bouvier, V.; Very, P.; Hudelot, C.; and Chastagnol, C. 2019.
\newblock Hidden Covariate Shift: A Minimal Assumption For Domain Adaptation.
\newblock \emph{arXiv preprint arXiv:1907.12299} .

\bibitem[{Buitinck et~al.(2013)Buitinck, Louppe, Blondel, Pedregosa, Mueller,
  Grisel, Niculae, Prettenhofer, Gramfort, Grobler, Layton, VanderPlas, Joly,
  Holt, and Varoquaux}]{sklearn_api}
Buitinck, L.; Louppe, G.; Blondel, M.; Pedregosa, F.; Mueller, A.; Grisel, O.;
  Niculae, V.; Prettenhofer, P.; Gramfort, A.; Grobler, J.; Layton, R.;
  VanderPlas, J.; Joly, A.; Holt, B.; and Varoquaux, G. 2013.
\newblock {API} design for machine learning software: experiences from the
  scikit-learn project.
\newblock In \emph{ECML PKDD Workshop: Languages for Data Mining and Machine
  Learning}, 108--122.

\bibitem[{Cao et~al.(2018)Cao, Ma, Long, and Wang}]{cao2018partial}
Cao, Z.; Ma, L.; Long, M.; and Wang, J. 2018.
\newblock Partial adversarial domain adaptation.
\newblock In \emph{Proceedings of the European Conference on Computer Vision
  (ECCV)}, 135--150.

\bibitem[{Chattopadhyay et~al.(2013)Chattopadhyay, Fan, Davidson, Panchanathan,
  and Ye}]{chattopadhyay2013joint}
Chattopadhyay, R.; Fan, W.; Davidson, I.; Panchanathan, S.; and Ye, J. 2013.
\newblock Joint transfer and batch-mode active learning.
\newblock In \emph{International Conference on Machine Learning}, 253--261.

\bibitem[{Chen et~al.(2019{\natexlab{a}})Chen, Xie, Huang, Rong, Ding, Huang,
  Xu, and Huang}]{chen2019progressive}
Chen, C.; Xie, W.; Huang, W.; Rong, Y.; Ding, X.; Huang, Y.; Xu, T.; and Huang,
  J. 2019{\natexlab{a}}.
\newblock Progressive feature alignment for unsupervised domain adaptation.
\newblock In \emph{Proceedings of the IEEE Conference on Computer Vision and
  Pattern Recognition}, 627--636.

\bibitem[{Chen et~al.(2019{\natexlab{b}})Chen, Wang, Long, and
  Wang}]{chen2019transferability}
Chen, X.; Wang, S.; Long, M.; and Wang, J. 2019{\natexlab{b}}.
\newblock Transferability vs. discriminability: Batch spectral penalization for
  adversarial domain adaptation.
\newblock In \emph{International Conference on Machine Learning}, 1081--1090.

\bibitem[{Combes et~al.(2020)Combes, Zhao, Wang, and Gordon}]{combes2020domain}
Combes, R. T.~d.; Zhao, H.; Wang, Y.-X.; and Gordon, G. 2020.
\newblock Domain Adaptation with Conditional Distribution Matching and
  Generalized Label Shift.
\newblock \emph{arXiv preprint arXiv:2003.04475} .

\bibitem[{Deng et~al.(2009)Deng, Dong, Socher, Li, Li, and
  Fei-Fei}]{deng2009imagenet}
Deng, J.; Dong, W.; Socher, R.; Li, L.-J.; Li, K.; and Fei-Fei, L. 2009.
\newblock Imagenet: A large-scale hierarchical image database.
\newblock In \emph{2009 IEEE conference on computer vision and pattern
  recognition}, 248--255. Ieee.

\bibitem[{Ganin and Lempitsky(2015)}]{ganin2015unsupervised}
Ganin, Y.; and Lempitsky, V. 2015.
\newblock Unsupervised Domain Adaptation by Backpropagation.
\newblock In \emph{International Conference on Machine Learning}, 1180--1189.

\bibitem[{Ganin et~al.(2016)Ganin, Ustinova, Ajakan, Germain, Larochelle,
  Laviolette, Marchand, and Lempitsky}]{ganin2016domain}
Ganin, Y.; Ustinova, E.; Ajakan, H.; Germain, P.; Larochelle, H.; Laviolette,
  F.; Marchand, M.; and Lempitsky, V. 2016.
\newblock Domain-adversarial training of neural networks.
\newblock \emph{The Journal of Machine Learning Research} 17(1): 2096--2030.

\bibitem[{Geva, Goldberg, and Berant(2019)}]{geva2019we}
Geva, M.; Goldberg, Y.; and Berant, J. 2019.
\newblock Are We Modeling the Task or the Annotator? An Investigation of
  Annotator Bias in Natural Language Understanding Datasets.
\newblock \emph{arXiv preprint arXiv:1908.07898} .

\bibitem[{Hanneke et~al.(2014)}]{hanneke2014theory}
Hanneke, S.; et~al. 2014.
\newblock Theory of disagreement-based active learning.
\newblock \emph{Foundations and Trends{\textregistered} in Machine Learning}
  7(2-3): 131--309.

\bibitem[{He et~al.(2016)He, Zhang, Ren, and Sun}]{he2016deep}
He, K.; Zhang, X.; Ren, S.; and Sun, J. 2016.
\newblock Deep residual learning for image recognition.
\newblock In \emph{Proceedings of the IEEE conference on computer vision and
  pattern recognition}, 770--778.

\bibitem[{Hsu and Lin(2015)}]{hsu2015active}
Hsu, W.-N.; and Lin, H.-T. 2015.
\newblock Active learning by learning.
\newblock In \emph{Twenty-Ninth AAAI conference on artificial intelligence}.
  Citeseer.

\bibitem[{Hunter(2007)}]{Hunter:2007}
Hunter, J.~D. 2007.
\newblock Matplotlib: A 2D graphics environment.
\newblock \emph{Computing in Science \& Engineering} 9(3): 90--95.
\newblock \doi{10.1109/MCSE.2007.55}.

\bibitem[{Johansson, Sontag, and Ranganath(2019)}]{johansson2019support}
Johansson, F.; Sontag, D.; and Ranganath, R. 2019.
\newblock Support and Invertibility in Domain-Invariant Representations.
\newblock In \emph{The 22nd International Conference on Artificial Intelligence
  and Statistics}, 527--536.

\bibitem[{Krizhevsky, Sutskever, and Hinton(2012)}]{krizhevsky2012imagenet}
Krizhevsky, A.; Sutskever, I.; and Hinton, G.~E. 2012.
\newblock Imagenet classification with deep convolutional neural networks.
\newblock In \emph{Advances in neural information processing systems},
  1097--1105.

\bibitem[{Liu et~al.(2019)Liu, Long, Wang, and Jordan}]{liu2019transferable}
Liu, H.; Long, M.; Wang, J.; and Jordan, M. 2019.
\newblock Transferable Adversarial Training: A General Approach to Adapting
  Deep Classifiers.
\newblock In \emph{International Conference on Machine Learning}, 4013--4022.

\bibitem[{Long et~al.(2015)Long, Cao, Wang, and Jordan}]{long2015learning}
Long, M.; Cao, Y.; Wang, J.; and Jordan, M.~I. 2015.
\newblock Learning transferable features with deep adaptation networks.
\newblock In \emph{Proceedings of the 32nd International Conference on
  International Conference on Machine Learning-Volume 37}, 97--105. JMLR. org.

\bibitem[{Long et~al.(2018)Long, Cao, Wang, and Jordan}]{long2018conditional}
Long, M.; Cao, Z.; Wang, J.; and Jordan, M.~I. 2018.
\newblock Conditional adversarial domain adaptation.
\newblock In \emph{Advances in Neural Information Processing Systems},
  1640--1650.

\bibitem[{Long et~al.(2016)Long, Zhu, Wang, and Jordan}]{long2016unsupervised}
Long, M.; Zhu, H.; Wang, J.; and Jordan, M.~I. 2016.
\newblock Unsupervised domain adaptation with residual transfer networks.
\newblock In \emph{Advances in Neural Information Processing Systems},
  136--144.

\bibitem[{Long et~al.(2017)Long, Zhu, Wang, and Jordan}]{long2017deep}
Long, M.; Zhu, H.; Wang, J.; and Jordan, M.~I. 2017.
\newblock Deep transfer learning with joint adaptation networks.
\newblock In \emph{Proceedings of the 34th International Conference on Machine
  Learning-Volume 70}, 2208--2217. JMLR. org.

\bibitem[{Marcus(2020)}]{marcus2020next}
Marcus, G. 2020.
\newblock The Next Decade in AI: Four Steps Towards Robust Artificial
  Intelligence.
\newblock \emph{arXiv preprint arXiv:2002.06177} .

\bibitem[{Oquab et~al.(2014)Oquab, Bottou, Laptev, and
  Sivic}]{oquab2014learning}
Oquab, M.; Bottou, L.; Laptev, I.; and Sivic, J. 2014.
\newblock Learning and transferring mid-level image representations using
  convolutional neural networks.
\newblock In \emph{Proceedings of the IEEE conference on computer vision and
  pattern recognition}, 1717--1724.

\bibitem[{Ouali et~al.(2020)Ouali, Bouvier, Tami, and
  Hudelot}]{ouali2020target}
Ouali, Y.; Bouvier, V.; Tami, M.; and Hudelot, C. 2020.
\newblock Target Consistency for Domain Adaptation: when Robustness meets
  Transferability.
\newblock \emph{arXiv preprint arXiv:2006.14263} .

\bibitem[{Pan and Yang(2009)}]{pan2009survey}
Pan, S.~J.; and Yang, Q. 2009.
\newblock A survey on transfer learning.
\newblock \emph{IEEE Transactions on knowledge and data engineering} 22(10):
  1345--1359.

\bibitem[{Paszke et~al.(2019)Paszke, Gross, Massa, Lerer, Bradbury, Chanan,
  Killeen, Lin, Gimelshein, Antiga et~al.}]{paszke2019pytorch}
Paszke, A.; Gross, S.; Massa, F.; Lerer, A.; Bradbury, J.; Chanan, G.; Killeen,
  T.; Lin, Z.; Gimelshein, N.; Antiga, L.; et~al. 2019.
\newblock PyTorch: An imperative style, high-performance deep learning library.
\newblock In \emph{Advances in Neural Information Processing Systems},
  8024--8035.

\bibitem[{Peng et~al.(2017)Peng, Usman, Kaushik, Hoffman, Wang, and
  Saenko}]{peng2017visda}
Peng, X.; Usman, B.; Kaushik, N.; Hoffman, J.; Wang, D.; and Saenko, K. 2017.
\newblock Visda: The visual domain adaptation challenge.
\newblock \emph{arXiv preprint arXiv:1710.06924} .

\bibitem[{Quionero-Candela et~al.(2009)Quionero-Candela, Sugiyama,
  Schwaighofer, and Lawrence}]{quionero2009dataset}
Quionero-Candela, J.; Sugiyama, M.; Schwaighofer, A.; and Lawrence, N.~D. 2009.
\newblock \emph{Dataset shift in machine learning}.
\newblock The MIT Press.

\bibitem[{Rai et~al.(2010)Rai, Saha, Daum{\'e}~III, and
  Venkatasubramanian}]{rai2010domain}
Rai, P.; Saha, A.; Daum{\'e}~III, H.; and Venkatasubramanian, S. 2010.
\newblock Domain adaptation meets active learning.
\newblock In \emph{Proceedings of the NAACL HLT 2010 Workshop on Active
  Learning for Natural Language Processing}, 27--32. Association for
  Computational Linguistics.

\bibitem[{Roth and Small(2006)}]{roth2006margin}
Roth, D.; and Small, K. 2006.
\newblock Margin-based active learning for structured output spaces.
\newblock In \emph{European Conference on Machine Learning}, 413--424.
  Springer.

\bibitem[{Saenko et~al.(2010)Saenko, Kulis, Fritz, and
  Darrell}]{saenko2010adapting}
Saenko, K.; Kulis, B.; Fritz, M.; and Darrell, T. 2010.
\newblock Adapting visual category models to new domains.
\newblock In \emph{European conference on computer vision}, 213--226. Springer.

\bibitem[{Saha et~al.(2011)Saha, Rai, Daum{\'e}, Venkatasubramanian, and
  DuVall}]{saha2011active}
Saha, A.; Rai, P.; Daum{\'e}, H.; Venkatasubramanian, S.; and DuVall, S.~L.
  2011.
\newblock Active supervised domain adaptation.
\newblock In \emph{Joint European Conference on Machine Learning and Knowledge
  Discovery in Databases}, 97--112. Springer.

\bibitem[{Sener and Savarese(2017)}]{sener2017active}
Sener, O.; and Savarese, S. 2017.
\newblock Active learning for convolutional neural networks: A core-set
  approach.
\newblock \emph{arXiv preprint arXiv:1708.00489} .

\bibitem[{Settles(2009)}]{settles2009active}
Settles, B. 2009.
\newblock Active learning literature survey.
\newblock Technical report, University of Wisconsin-Madison Department of
  Computer Sciences.

\bibitem[{Su et~al.(2020)Su, Tsai, Sohn, Liu, Maji, and
  Chandraker}]{su2020active}
Su, J.-C.; Tsai, Y.-H.; Sohn, K.; Liu, B.; Maji, S.; and Chandraker, M. 2020.
\newblock Active adversarial domain adaptation.
\newblock In \emph{The IEEE Winter Conference on Applications of Computer
  Vision}, 739--748.

\bibitem[{Vaswani et~al.(2017)Vaswani, Shazeer, Parmar, Uszkoreit, Jones,
  Gomez, Kaiser, and Polosukhin}]{vaswani2017attention}
Vaswani, A.; Shazeer, N.; Parmar, N.; Uszkoreit, J.; Jones, L.; Gomez, A.~N.;
  Kaiser, {\L}.; and Polosukhin, I. 2017.
\newblock Attention is all you need.
\newblock In \emph{Advances in neural information processing systems},
  5998--6008.

\bibitem[{Walt, Colbert, and Varoquaux(2011)}]{walt2011numpy}
Walt, S. v.~d.; Colbert, S.~C.; and Varoquaux, G. 2011.
\newblock The NumPy array: a structure for efficient numerical computation.
\newblock \emph{Computing in science \& engineering} 13(2): 22--30.

\bibitem[{Wang and Shang(2014)}]{wang2014new}
Wang, D.; and Shang, Y. 2014.
\newblock A new active labeling method for deep learning.
\newblock In \emph{2014 International joint conference on neural networks
  (IJCNN)}, 112--119. IEEE.

\bibitem[{Yosinski et~al.(2014)Yosinski, Clune, Bengio, and
  Lipson}]{yosinski2014transferable}
Yosinski, J.; Clune, J.; Bengio, Y.; and Lipson, H. 2014.
\newblock How transferable are features in deep neural networks?
\newblock In \emph{Advances in neural information processing systems},
  3320--3328.

\bibitem[{You et~al.(2019)You, Long, Cao, Wang, and Jordan}]{you2019universal}
You, K.; Long, M.; Cao, Z.; Wang, J.; and Jordan, M.~I. 2019.
\newblock Universal domain adaptation.
\newblock In \emph{Proceedings of the IEEE Conference on Computer Vision and
  Pattern Recognition}, 2720--2729.

\bibitem[{Zhang et~al.(2018)Zhang, Ding, Li, and
  Ogunbona}]{zhang2018importance}
Zhang, J.; Ding, Z.; Li, W.; and Ogunbona, P. 2018.
\newblock Importance weighted adversarial nets for partial domain adaptation.
\newblock In \emph{Proceedings of the IEEE Conference on Computer Vision and
  Pattern Recognition}, 8156--8164.

\bibitem[{Zhao et~al.(2019)Zhao, Des~Combes, Zhang, and
  Gordon}]{zhao2019learning}
Zhao, H.; Des~Combes, R.~T.; Zhang, K.; and Gordon, G. 2019.
\newblock On Learning Invariant Representations for Domain Adaptation.
\newblock In \emph{International Conference on Machine Learning}, 7523--7532.

\end{thebibliography}

\newpage

\appendix

\section{Proof of the theoretical analysis} 
The naive classifier's target error is bounded as follows:
\begin{equation}
    \varepsilon_T(h_{\mathcal A}) \leq \left (\frac{1}{b \pi} - 1\right) (\varepsilon_S(h)  + 8 \tau + \eta )
\end{equation}
where $\tau := \sup_{\mathsf f \in \mathsf F} \{ \mathbb E_{x\sim p_T}\left [h_{\mathcal A}(x) \cdot \mathsf f(\varphi(x))\right ] - \mathbb E_{x,y \sim p_S}[y\cdot \mathsf f(\varphi(x))] \}$ is the transferability error, $\mathsf F$ is the set of continuous functions from $\mathcal Z$ to $[-1, 1]^{C}$ and $\eta:=\inf_{\mathsf f \in \mathsf F}\varepsilon_T(\mathsf f\varphi)$. 

\begin{proof}
First, we observe that : 

\begin{equation}
    \varepsilon_T(h_{\mathcal A}) \leq \beta \varepsilon_T(h)
\end{equation}
with $\beta = 1 - \frac{b\pi}{\varepsilon_T(h)} $. Then;
\begin{equation}
    \rho := \frac{\beta}{ 1- \beta} = \frac{\varepsilon_T(h)}{b\pi} -1 \leq \frac{1}{b\pi}-1
\end{equation}
We apply Bound 4 from \cite{bouvier2020robust} where the inductive classifier is the active classifer $h_{\mathcal A}$. Then we bound the invariance error using the transferability error (Proposition 3, item 1 from \cite{bouvier2020robust}) to obtain $6\tau + 2\tau = 8 \tau$, leading to the announced result.
\end{proof}

\end{document}